\documentclass[10pt,journal,compsoc]{IEEEtran}
\ifCLASSOPTIONcompsoc
  \usepackage[nocompress]{cite}
\else
  \usepackage{cite}
\fi

\usepackage[center]{caption}

\usepackage{amssymb}
\usepackage{graphicx}
\usepackage{latexsym}
\usepackage{epstopdf}
\usepackage{amsmath}
 \usepackage{here}
\usepackage{url}
\usepackage{xcolor}

\usepackage{caption}
\usepackage{subcaption}
\usepackage{tabularx}
\usepackage{lineno}
\ifCLASSINFOpdf
\else
\fi

\begin{document}
%
\title{Deep and Shallow Covariance Feature Quantization for 3D Facial Expression Recognition}
%
%
%
%

\author{Walid~Hariri,~~
        Nadir~Farah,~~
				Dinesh Kumar Vishwakarma
\IEEEcompsocitemizethanks{\IEEEcompsocthanksitem M. Hariri was with the Department
of Computer Science, Badji Mokhtar Annaba University, Algeria,
BP12, 23000.\protect\\
E-mail: hariri@labged.net
\IEEEcompsocthanksitem M. Farah was with the Department
of Computer Science, Badji Mokhtar Annaba University, Algeria. \protect\\
E-mail: farah@labged.net
\IEEEcompsocthanksitem M. Vishwakarma was with the Department
of Information Technology, Delhi Technological University, New Delhi-110042, India. \protect\\
E-mail: dinesh@dtu.ac.in}
\thanks{Manuscript received.}}

%
%

\markboth{Journal of \LaTeX\ Class Files,~Vol. , No. , 2021}%
{Shell \MakeLowercase{\textit{et al.}}: Bare Demo of IEEEtran.cls for Computer Society Journals}
%



\IEEEtitleabstractindextext{%
\begin{abstract}
Facial expressions recognition (FER) of 3D face scans has received a significant amount of attention in recent years. Most of the facial expression recognition methods have been proposed using mainly 2D images. These methods suffer from several issues like illumination changes and pose variations. Moreover, 2D mapping from 3D images may lack some geometric and topological characteristics of the face. Hence, to overcome this problem, a multi-modal 2D + 3D feature-based method is proposed. We extract shallow features from the 3D images, and deep features using Convolutional Neural Networks (CNN) from the transformed 2D images. Combining these features into a compact representation uses covariance matrices as descriptors for both features instead of single-handedly descriptors. A covariance matrix learning is used as a manifold layer to reduce the deep covariance matrices size and enhance their discrimination power while preserving their manifold structure. We then use the Bag-of-Features (BoF) paradigm to quantize the covariance matrices after flattening. Accordingly, we obtained two codebooks using shallow and deep features. The global codebook is then used to feed an SVM classifier. High classification performances have been achieved on the BU-3DFE and Bosphorus datasets compared to the state-of-the-art methods.
\end{abstract}

}

\maketitle

\IEEEdisplaynontitleabstractindextext

%
\IEEEpeerreviewmaketitle

\section{Introduction}
\label{sec:intro}

Facial expressions reported up to 55\% of face to face communication while only 7\% and 38\% of the emotional expression are allocated to oral language and vocal tone respectively \cite{ismail2019usage}. Therefore, the recognition of human emotion from facial expression images has become a very interesting research field in computer vision and pattern recognition. The majority of this research focuses on recognizing six basic expressions \cite{fang20113d,shan2009facial,li2020deep} namely: happy (HA), sad (SA), disgust (DI), surprise (SU), fear (FE), and angry (AN), defined by Ekman and Friesen \cite{ekman1971constants} and accepted as universal emotions. 
Due to the diverse sources of variability in 2D and 3D facial images, FER has proven to be a very difficult task. Such variations can be environment-related (light conditions, occlusions caused by certain objects), subject-related (location variation), and acquisition-related (image size, distortion, vibration, and other imperfections).

Among previous facial expression recognition (FER) works, we distinguish 2D data-based methods \cite{zeng2009survey,mahersia2015using}. Despite the good performance which has been achieved in 2D FER, it is still a challenging task as it has to deal with two main issues: illumination and pose. Moreover, the texture, image resolution, and color are not necessarily the same when the images are acquired in unconstrained conditions. In such a situation, 2D FER struggles to achieve high performance.

3D data, on the other hand, rely on facial structure and provide more geometrical characteristics and are less sensitive to light conditions \cite{dutta2021complement,patil20153} and pose variations \cite{ocegueda20133d}. They have the potential to maintain both geometric and topological facial structural details with the depth information. Accordingly, in 3D FER, 3D data can efficiently capture all the facial parts' movement of the face also in unconstrained conditions. This particularity explains the robustness of the 3D modality for FER due to the recent progress of 3D acquisition techniques and low-cost 3D sensors (e.g., Microsoft Kinect, Intel RealSense) \cite{yin20063d}. Thereby, various databases for 3D facial expression analysis have appeared and have been used by the research community to evaluate their algorithms. Among these databases, we find BU-3DFE ~\cite{yin20063d}, BU-4DFE, and Bosphorus~\cite{savran2008bosphorus} that contained the six basic emotions. Differently, FRGC v2.0 and GAVAB, present a set of expression variations, but not with a regular distribution thereby, they are not recommended for the FER.

3D FER methods in the literature can be carried out using purely 3D data or of its combination with texture and time variation information. 3D face scan can be also mapped into 2D representations such as three normal component maps, curvedness map,  Gaussian curvature map, Mean curvature map, geometry map and texture map, etc \cite{alexandre2020systematic}. 2D-3D multi-modality could give diversity to the extracted features and outperforms single-modality-based methods. Achieving a good performance, however, requires efficient and discriminative features. In the following, we present and discuss the most important existing FER methods in the literature, and their use in multi-modal representation.

\section{Literature review}
\label{sec:re_work}
In recent years, Convolutional Neural Networks have been widely deployed in a large range of tasks including image classification systems, particularly facial expression classification \cite{krizhevsky2012imagenet,lopes2017facial,li2020deep}. The three main advantages of using CNNs for deep learning are the elimination of handcrafted extraction of features, the cutting-edge recognition results, and the ability to use pre-trained networks for other recognition tasks.

Deep CNNs are generally applied to FER using 2D images to learn deeper feature representations of facial expression. Despite their potential, they can't achieve high performances when dealing with considerable illumination and pose variations \cite{vo20193d}. To overcome this problem, various approaches have used 3D data for learning CNN, such as Volumetric CNN \cite{wu20153d}, Field probing neural networks \cite{li2016fpnn}, 3D Graph-CNN \cite{papadopoulos2021face}, multi-view CNN \cite{su2015multi} and Vote3D \cite{wang2015voting}. The high cost of these operations and even more the very large point clouds is a bigger challenge. To obviate this drawback, recent methods proposed to normalize the 3D face image to lower dimensions (e.g. 2D depth image, principle curvatures map), all of which are jointly fed into CNN for feature learning and classification. For example, Jan et al. 2018 \cite{jan2018accurate} learned deep CNN features from a 2D texture map and a depth map extracted from 3D face scans, then an SVM is applied for the classification. They thus take advantage of CNN as a deep feature extractor for FER applications. However, normalizing the 3D face scans to lower dimensions may divest the geometrical and topological information. These limitations make the 3D FER a very challenging task. Therefore, the multi-modal 2D+3D has become a frequent approach for FER, commonly used in literature. One of the most efficient approaches that successfully utilized a 2D+3D multi-modal-based system is proposed in \cite{li2017multimodal}. From 3D face scans, the authors extracted six 2D map representations involve texture map, and combined them through feature learning and fusion learning into a single end-to-end training framework. In \cite{oyedotun2017facial}, RGB images are combined with depth maps to a deep CNN from scratch, in addition to transfer learning using two pre-trained models (ResNet50 and VGG-19) as a hierarchical feature representation. From the state-of-the-art review, one can notice that learning a deep model from scratch is not suitable for 3D FER due to the lack of a large amount of data. Alternatively, pre-trained models show very high interest to overcome this problem and it can be applied in two different strategies. In the first strategy, the facial expression images are fed to the pre-trained model, and the fully connected (FC) layers were typically replaced by one or more additional layers then the networks re-trained to adapt the weights of the added layers for FER. This strategy is called fine-tuning because the pre-trained models are adapted to the FER problem. As an example, a supervised fine-tuning on a small dataset  is applied in \cite{ng2015deep} for FER. The second strategy aims to apply the pre-trained models as feature generators, then uses the obtained features extracted from a FC layer or convolution layer. Traditional ML algorithms (e.g. SVM classifier) can be added to the system, trained with the generated deep features to improve the performance and to avoid the over-fitting problem. However, the FC layers of the pre-trained models are more dataset-specific features (generally pre-trained on ImageNet dataset) which is a very different dataset, thus, this strategy is not suitable for FER in such a case. 

On the other hand, multiple pre-trained CNN models of different architecture that represent different feature abstraction levels can be combined using ensemble model. The global decision takes into account the decision of each model by applying a vote or weighted sum operation. For example, weighted prediction scores of the pre-trained VGG-16 and AlexNet models are computed in \cite{fan2018multi} to classify the expressions. Despite the high performances achieved by these models, the problem of over-fitting is still a grand challenge because the pre-trained models are generally trained with ImageNet dataset which has a lot of data belonging to very different classes, and the performance is highly depend on the ensemble strategy that cannot be meaningful for FER. 

\section{Contribution of the paper}
\label{sec:contr}

We propose in this paper a 2D+3D multi-modal approach that performs two deep pre-trained models (AlexNet, VGG-16) applied to generate deep features from 2D (depth and curvature) maps, and handcrafted features extracted from the 3D images. The obtained features are firstly co-varied and normalized using a feature quantization to feed an SVM classifier. Our objectives are threefold: \textbf{I)} we avoid the over-fitting that can be occurred after fine-tuning the pre-trained models to small FER datasets. \textbf{II)} local features extracted from the 3D images may provide more geometrical and spatial characteristics that could be lacked after mapping the 3D images onto a 2D (depth or curvature) map. \textbf{III)} Ensure a high discriminative power from the pre-trained models instead of applying scratch training, which is time-consuming, and is not suitable for small 3D FER datasets.

Accordingly, instead of using the generated deep features directly to feed a ML classifier, we take full advantage of using covariance matrices as local descriptors which have many benefits: \textbf{I)} they provide a natural way for fusing multi-modal features that can be of different dimensions. \textbf{II)} compact representation since covariance matrices can be computed from different sized regions, the obtained covariance descriptors are of the same size whatever the size of their features. \textbf{III)} ability to compare any regions of different sizes. 

It is also important to note that the classical use of deep learning approaches mainly focused on handling data in Euclidean space. However, it is not always the case when dealing with other structures such as Symmetric Positive Definite (SPD) matrices (the space is indicated also by $Sym_{d}^{+}$) that deal with non-Euclidean space. Therefore, various methods have been proposed to be able to apply deep learning approaches on non-Euclidean space, i.e. Lie groups \cite{huang2017deep}, SPD manifolds \cite{huang2017riemannian} or Grassmann manifolds \cite{huang2018building}. 

In order to use SPD matrices as an input of classical classifiers that usually assume a Euclidean geometry, Harandi et~al. 2014 \cite{harandi2014expanding} and Jayasumana et~al. 2013 \cite{jayasumana2013kernel} have applied a non-linear mapping into a high dimensional space using kernel-based methods via the use of a projection matrix. This solution preserves the SPD structure, however, it is still incapable to deal with non-linear learning. SPD learning, on the other hand, solves this problem and offers the possibility to introduce a non-linearity during learning SPD matrices in the network. This motivated us to use this strategy to get a high discriminative representation, and to accurately classify the expressions. To have the same dimensionality of the resulting features, we employed a feature quantization to feed an SVM classifier. 

The main contributions of this papers are summarized as follows: 

\begin{itemize}
	\item Deep features generated from the last convolutional layer of two modified pre-trained deep CNNs (VGG-16 and AlexNet) using 2D depth and curvature map images.
	\item Geometrical features are captured from the 3D images.
	\item The extracted features are embedded into a covariance pooling for the dimensionality reduction.
	\item To enhance the discrimination power, deep covariance learning is used through two additional layers (BiMap layer + Eigenvalue Rectification).
	\item  Feature quantization of flattened deep and shallow covariance matrices is carried out which takes full advantage of both geometrical features and deep features of the whole face.
	\item An experiment is performed on public datasets such as BU-3DFE and Bosphorus. Further, the performance is compared to similar state-of-the-art methods and shows the superiority of the proposed methodology. 
\end{itemize}

The rest of the paper is structured as follows, an overview of the proposed method is given in Section~\ref{sec:method} including four steps: pre-processing, feature extraction, the deep learning of SPD matrices as well as the shallow and deep Bag-of-Features paradigm. 
Experimental results and a detailed comparative study are presented in Section \ref{sec:exp}. 
The obtained results are analyzed and comprehensively discussed in Section \ref{sec:disc}. Conclusions end the paper.

\begin{figure*}[!t]
\centering
\includegraphics[width=.77\textwidth]{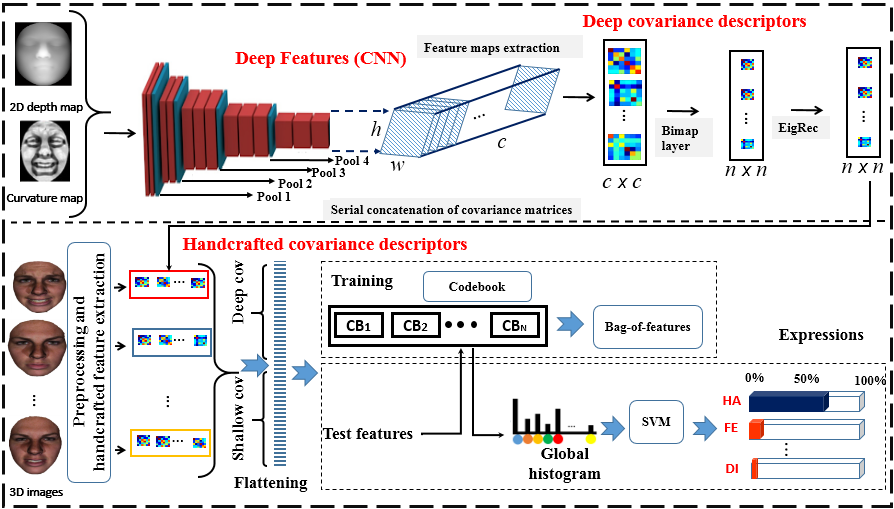}
\caption{The Proposal Overview.}
\label{fig:overview}
\end{figure*}

\begin{figure}[!t]
\centering
\includegraphics[width=.47\textwidth]{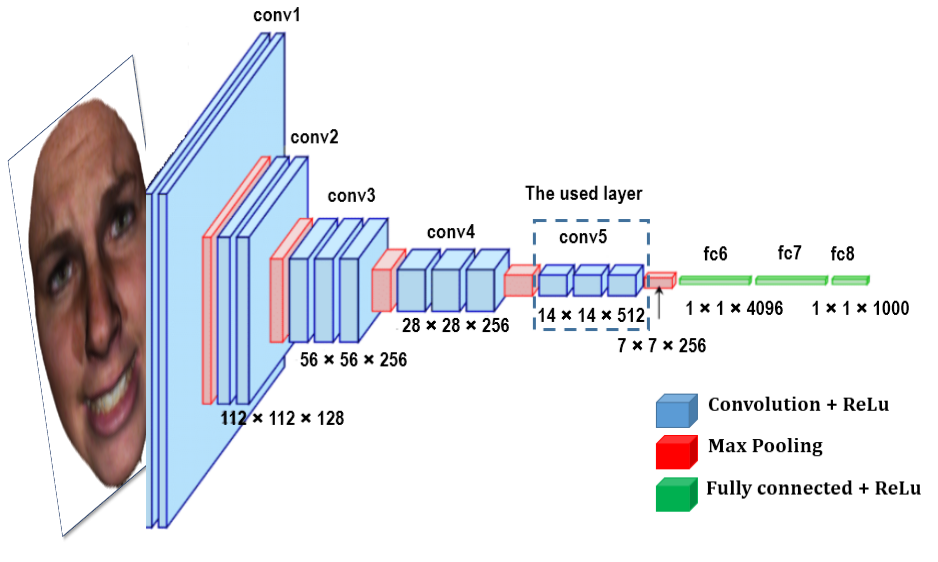}
\caption{VGG-16 network architecture proposed in \cite{simonyan2014very}.}
\label{fig:VGG}
\end{figure}

\begin{figure}[!t]
\centering
\includegraphics[width=.28\textwidth]{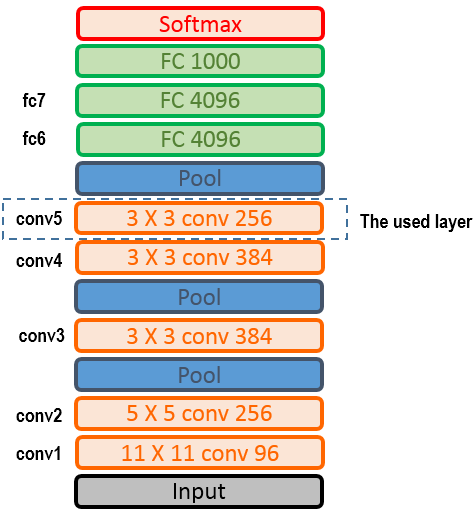}
\caption{AlexNet network architecture proposed in \cite{krizhevsky2012imagenet}.}
\label{fig:Alex}
\end{figure}

\section{Method}
\label{sec:method} 

The proposed methodology consists of a deep and shallow features combination based approach as given in Fig.~\ref{fig:overview}. 
In order to exploit the discriminative power of the deeply learned features using CNN, and the efficiency of covariance descriptors as a compact representation, we propose in this paper a deep and shallow feature combination method using covariance descriptors to handle the problem of FER on the two challenging BU-3DFE and Bosphorus datasets. Shallow features extract spatial and geometrical characteristics from the 3D face images. In the deep stage, we feed the pre-trained CNNs with the normalized face images. To pool the feature maps spatially from the CNN, we propose to use covariance pooling, and then employ the manifold network to deeply learn the second-order statistics. BoF paradigm is then applied to quantize the deep and shallow covariance matrices after flattening. SVM is applied to classify the expressive faces. In the following, we explain each stage of the proposed method from the pre-processing to the classification.

\subsection{Pre-processing and data transformation}
\label{sec:prep} 
Data pre-processing aims to eliminate deficiencies like holes, and unnecessary regions such as hair, neck, and clothes from the face surface. We thereby employ some corrections and filters; involving \textit{a smoothing process} that relieves spikes, \textit{a cropping filter} to keep only the desired portion of the face, \textit{a filling holes}, and \textit{a median filter} to withdraw spikes. In the case of deep covariance, each 3D face model is firstly transformed to 2D map images namely: 2D depth and principal curvatures. \textit{The 2D depth map} shows the gray value of each image pixel which represents the depth of the associated point on the 3D facial scan. \textit{The Principal curvatures map} on the other hand represents the principal curvature values over the 3D mesh. It is approximated by the local cubic fitting method \cite{goldfeather2004novel}.
All 2D faces are then normalized to 224 $\times$ 224 pixels. 

\subsection{Feature extraction}
\label{sec:cov_mat}

Once the 3D face scans have been pre-processed, we extract covariance descriptors from deep and shallow features as follows: 

\subsubsection{Deep covariance features : } 
\label{sec:deep_cov}
We extract deep features using VGG-16 and AlexNet models (see Fig.~\ref{fig:VGG} and Fig.~\ref{fig:Alex}). 

VGG-16 face model \cite{simonyan2014very} is a deep CNN model pre-trained on the ImageNet database \cite{deng2009imagenet}. It contains 16 layers, trained on the ImageNet dataset which has over 14 million images and 1000 classes This model has been successfully used for face recognition \cite{parkhi2015deep} and facial expression recognition \cite{ma2021multi}.

The second deep CNN model used to extract deep features is AlexNet \cite{krizhevsky2012imagenet} pre-trained on the ImageNet database. It
consists of 25 layers including convolution, fully connected, pooling, Rectified Linear Units (ReLU), normalization. AlexNet significantly outperformed the runner-up with a top-5 error rate of 15.3\% in the 2012 ImageNet challenge \cite{iandola2016squeezenet}.

We only consider the feature maps (FMs) at the last convolutional layer of VGG-16 and AlexNet models, also called channels. Instead of using the softmax function to classify faces, we propose to extract covariance matrices as facial descriptors using the obtained FMs to get a higher discriminative power compared to the extracted deep features. Accordingly, we obtain a more efficient and compact representation that encodes the correlation between the extracted non-linear features within different spatial levels.  
	
Let $\mathcal{P}$ = $\{M_i, i=1\dots m\}$ be the set of feature maps extracted from the 2D map images  (i.e. 2D depth and principal curvatures) using VGG-face and AlexNet models separately. Each patch $M_i$ encodes the local geometric and spatial properties of the face image. 

The extracted features $\phi(f)$ are arranged in a $(c \times w \times h)$ tensor, where $w$ and $h$ denote the width and height of the FMs, respectively, and $c$ is the number of $FMs$. Each feature map $M_i$ is vectorized into a n-dimensional vector with $n$ = $w \times h$, and the input tensor is transformed to a set of $n$ observations stored in the matrix $[v_1; v_2; ... ; v_n] \in R^{c \times n}$. Each observation $v_i \in R^c$
encodes the values of the pixel $i$ across all the $m$ feature maps. Finally, we compute the corresponding ($c \times c$) covariance matrix $X_i$, defined by Eq.\ref{eq:1}.

\begin{equation}\label{eq:1}
X_i=\frac{1}{n}\sum_{j=1}^n(f_j-\mu)(f_j-\mu)^T
\end{equation}

\subsubsection{Shallow covariance features: }
\label{sec:handcr_cov}
We extract 40 patches of the same size from the face surface. Each patch has a reference point that is positioned in its center. We refer to these reference-points by $m$ and to each patch by $\rho_i$. The covariance matrices are then computed from each patch using their geometric features as proposed in \cite{hariri20173d}. 

Let $\mathcal{P} = \{\rho_i, i=1\dots m\}$ be the set of patches extracted from a 3D face. Each patch $\mathcal{P}_i$ defines a region around a feature point $p_i= (x_i, y_i, z_i)^t$. For each point $f_j$ in $\rho_i$,
we extract a feature vector $F_j$, of dimension $d$, which encodes the local geometrical information and spatial
characteristics of the point. In our experiments, we use the following feature vector defined as Eq.\ref{eq:2}:
\begin{equation}\label{eq:2}
F_j= \left[x_j,y_j,z_j,C,M,D_{j}\right]
\end{equation}	

where $x_j$, $y_j$ and $z_j$ are the three-dimensional coordinates of the point $p_j$. $M$ and $C$ are Mean curvature and Curvedness respectively. $D_{j}$ is the distance of $p_j$ from the origin. Each patch is defined by a covariance matrix, which is defined by Eq.\ref{eq:3}: 
\begin{equation}\label{eq:3}
P_i=\frac{1}{n}\sum_{j=1}^n(F_j-\mu)(F_j-\mu)^T    
\end{equation}

where $\mu$ is the mean of the feature vectors $\left\{F_{j}\right\}_{j=1...n}$ computed in the patch $\rho_i$, and $n$ is the number of points in $\rho_i$.


\subsection{Deep learning on SPD matrices}
\label{sec:Cov_section}
This section introduces the structure of SPD matrices and explain how to learn them on the $Sym_{d}^{+}$ space.\\
\textbf{Geometry of SPD matrices:}
the space of covariance matrices is presented as follows:

The $m\times m$ SPD matrix $X$ is has the particularity of $y^{T}X y >$ 0 while $y\in$ $\mathbf{R}^m$. The space of $m\times m$ SPD matrices, indicated by $Sym_{d}^{+}$ is not an Euclidean space but a non-linear Riemannian manifold of size $m \times (m+1)/2$ that supports a Riemannian metric (i.e. geodesic distance) called the Affine Invariant. It is given by: 

$\delta_{R}(\boldsymbol{A}, \boldsymbol{B})=\left\|\log \left(\boldsymbol{A}^{-1 / 2} \boldsymbol{B} \boldsymbol{A}^{-1 / 2}\right)\right\|_{F}$, where $A$ and $B$ are covariance matrices.

\textbf{SPD matrix learning}: consists of reducing the SPD matrices size and enhance their discrimination power while preserving their manifold structure. Thus, the output of the SPD matrices learning must still be SPD matrices too. For this reason, we can distinguish linear or non-linear transformations of SPD matrices. Among linear transformation, the Bimap layer has been used in \cite{zhang2018deep}. Non-linear transformation is similar to ReLU layers in CNN, it can be found in \cite{engin2018deepkspd}. Finally, to flatten the obtained covariance matrices, we apply eigenvalue decomposition (EIG) algorithm to feed a SVM classifier. In the following, we present these steps.

\subsubsection{Linear transformation}
\label{sec:lin}
We apply a bilinear mapping on the covariance matrices matrices using the BiMap layer in order to reduce their size and to be able to concatenate them serially with handcrafted based covariance matrices. The BiMap is considered as a fully connected layer since it preserves the geometrical structure of covariance matrices while reducing dimension as follows: 
if $X_{k-1}$ be input SPD matrix, $W_k$ $\in$ $R^{d_k*d_{k-1}}$ be weight matrix in the space of full rank matrices and $X_{k}$ $\in$ $R^{d_k*d_{k}}$ be output matrix, then $k^{th}$ the bilinear mapping $f^b_k$ is defined by Eq.\ref{eq:4}:
\begin{eqnarray}\label{eq:4}
X_{k}=f^b_k(X_{k-1};W_k)=W_k X_{k-1}W^k_T.
\end{eqnarray}

\subsubsection{Non-linear transformation}
\label{sec:non_lin}
Tuning up the covariance matrices is necessary to exploit the ReLU-like layers to introduce a non-linearity to the context of the deep SPD matrices. To do so, we apply the Eigenvalue Rectification presented in \cite{huang2017riemannian}. 

We note $X_{k-1}$ the input covariance matrix and $X_k$ the output one. $\epsilon$ be the eigenvalue rectification threshold, accordingly, the $k^{th}$ layer $f_r^k$ is defined by Eq.\ref{eq:5}: 

\begin{equation}\label{eq:5}
\mathbf{X}_{k}=f_{r}^{k}\left(\mathbf{X}_{k-1}\right)=\mathbf{U}_{k-1} \max \left(\epsilon \mathbf{I}, \sigma_{k-1}\right) \mathbf{U}_{k-1}^{T}
\end{equation}
where ${X}_{k-1}$ and $\Sigma_{k-1}$ are defined by eigenvalue decomposition $\mathbf{X}_{k-1}=\mathbf{U}_{k-1} \Sigma_{k-1} \mathbf{U}_{k-1}^{T}$. 

\subsubsection{SPD matrices flattening}
Log Eigenvalue Layer is applied to flatten the SPD matrices while preserving the Manifold structure. It consists of applying eigenvalue decomposition and log as matrix operation. 

We note $X_{k-1}$ the input covariance matrix and $X_k$ the output one. The LogEig layer applied in the $k^{th}$ layer $f_l^k$ is defined by Eq.\ref{eq:6}:

\begin{equation}\label{eq:6}
\mathbf{X}_{k}=f_{l}^{k}\left(\mathbf{X}_{k-1}\right)=\log \left(\mathbf{X}_{k-1}\right)=\mathbf{U}_{k-1} \log \left(\Sigma_{k-1}\right) \mathbf{U}_{k-1}^{T}
\end{equation} 

Where $\mathbf{X}_{k}=\mathbf{U}_{k-1} \Sigma_{k-1} \mathbf{U}_{k-1}^{T}$ is an eigenvalue decomposition and log is an element-wise matrix operation \cite{acharya2018covariance}.

\subsection{Shallow and deep Bag-of-Features paradigm and classification}
\label{sec:BoF}
The two sets of the flattened covariance matrices are of different sizes since they are computed from two different types of feature vectors. To overcome this problem, we apply the BoF paradigm proposed in \cite{o2011introduction} on the two sets of the extracted covariance descriptors (deep and shallow ones after flattening), separately. If we have $N$ covariance matrices of size $d \times d$, the obtained feature vector after flattening is of size $N \times \frac{(d \times d-1)}{2}$.

Note that the quantization of the obtained feature vectors for each image leads to two predefined number of histogram bins/codewords (deep and shallow histograms). This makes each histogram independent of the number of the obtained feature vectors and the image size. In the following, we refer to the codebooks obtained from the co-varied deep features by \textit{deep codebook}, and to that obtained from the co-varied shallow features by \textit{shallow codebook}. Once the two histograms are computed, we pass to the classification stage to assign each test image to its expression class. To do so, we apply SVM classifier where each face is represented by the global histogram which is the concatenation of the deep and the shallow codebooks.

\begin{figure}[h]
\centering
\captionsetup{justification=centering}
\includegraphics[width=.46\textwidth]{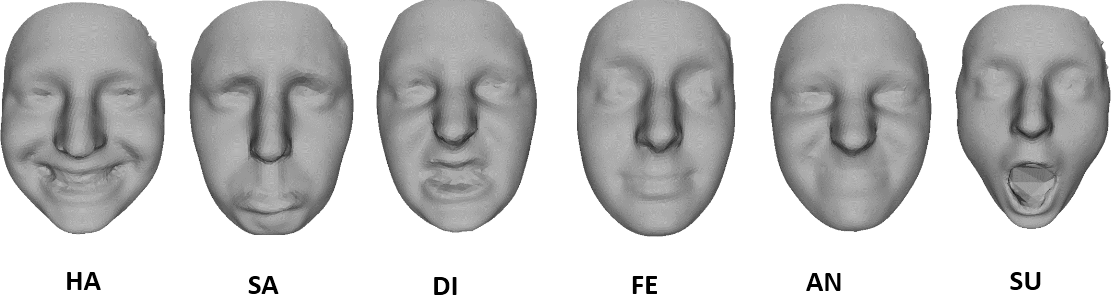}
\caption{3D face models from Bosphorus dataset.}\label{fig:Bos}
\end{figure}
\begin{figure}[h]
\centering
\captionsetup{justification=centering}
\includegraphics[width=.46\textwidth]{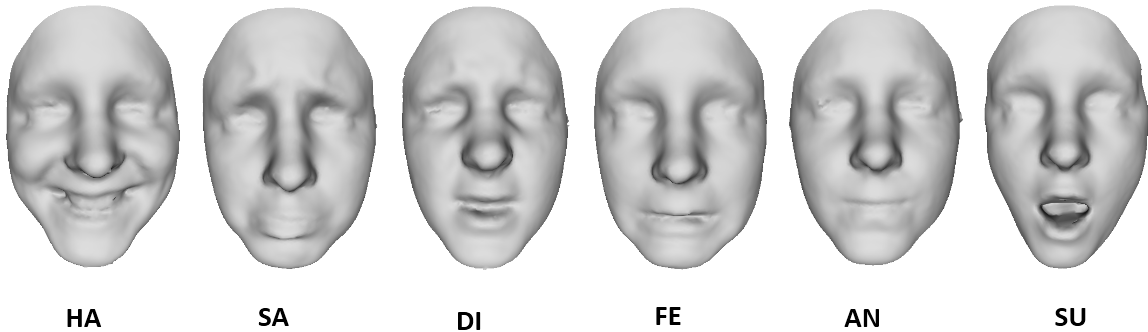}
\caption{3D face models from BU-3DFE dataset.}\label{fig:BU}
\end{figure}

\section{Experiments}
\label{sec:exp}
This section presents the datasets used to evaluate the proposed method and the experimental results compared to other methods in the literature using the same datasets. 

\subsection{Dataset description}

\textit{\textbf{The Bosphorus dataset}}~\cite{savran2008bosphorus} which was made for testing the algorithms using 3D and 2D facial images for facial analysis and recognition tasks. This dataset includes 105 subjects with many variations (total of 4666 images). It was obtained using structured-light technology to get 3D scans with the six expressions and neutral scan for each subject. Dimensions are defined by 0.3 mm, 0.3 mm, and 0.4 mm respectively. Fig.~\ref{fig:Bos} presents some examples from the Bosphorus dataset for the same subject. 

\textit{\textbf{BU-3DFE dataset}} (Binghamton University 3D Facial Expression) ~\cite{yin20063d} is a multi-view facial expression database of 2500 images captured in lab-controlled environment. It contains 3D expressive face scans with texture scans of 100 subjects. The basic facial expressions are induced by various ways and head poses with four intensity degrees. we can also find a neutral face associated to each subject. Fig.~\ref{fig:BU} presents an example of six expressive faces from BU-3DFE dataset. To make a fair comparison with the state-of-the-art methods, we apply the same protocol by excluding the neutral face and only use the six expressive faces.

                                      

\subsection{Method evaluation}
In this section, we present the different evaluation protocols and the experimental results to demonstrate the efficiency of the proposed method. After the pre-processing step, we extract deep and shallow based covariance descriptors as follows:

{\textbf{Deep feature-based covariance descriptors:}}
from each 2D map image (2D depth and principal curvatures), we employed the last convolutional layer of CNN models to extract deeper features from each face image. Using VGG-16, this layer contains 512 feature maps having the size of 14 $\times$ 14. Thus, we obtain 512 $\times$ 512 covariance descriptors as presented in Fig.~\ref{fig:VGG}. When dealing with AlexNet model, we obtain covariance descriptors of size 256 $\times$ 256. Next, we apply dimension reduction of the obtained covariance matrices with BiMap layer. The dimensionalities of the VGG-16 SPD transformation matrices are set to 512~$\times$~250, 250~$\times$~100, 100~$\times$~50. AlexNet SPD matrices are transformed to 256~$\times$~150, 150~$\times$~100, 100~$\times$~50 respectively (see Section \ref{sec:lin}). Finally, the obtained SPD matrices are tuned up to introduce a non-linearity to the context of the deep SPD matrices as presented in Section \ref{sec:non_lin}.

{\textbf{Shallow feature-based covariance descriptors:}} we extract 40 patches from the face surface. Each patch has a reference point that is positioned in its center. We refer to these reference-points by $m$ and to each patch by $\rho_i$. The radius of each patch is simply given by $r$ = $15 \times r/100$, where $r$ is the radius of the whole facial shape defined as a bounding sphere. 
 
To describe each parch region, we compute the covariance matrices of size $6\times6$ using the corresponding feature vector: $\left[x,y,z,C,M,D\right]$ as presented in Section~\ref{sec:handcr_cov}. 

{\textbf{Features quantization :}}
we finally apply the BoF paradigm on the two sets of covariance matrices described above after flattening as described in Section~\ref{sec:BoF}. We thus obtain two global histograms for quantization. SVM classifier is then applied on the serial concatenation of the two histograms to classify the six facial expressions.

\begin{figure}[h]
 \centering
\includegraphics[width=.50\textwidth]{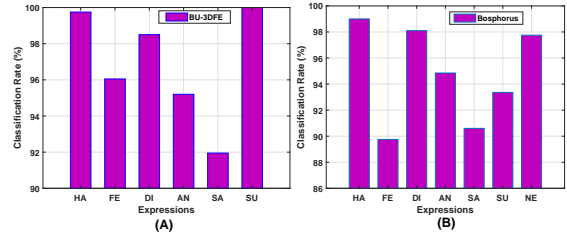}
\caption{Facial expression classification performance on (a) BU-3DFE and (b) Bosphorus datasets.}
\label{fig:rates_BU_BO}
\end{figure}

\subsubsection{Performances on BU-3DFE dataset}
\begin{figure}[h]
\centering
\includegraphics[width=.25\textwidth]{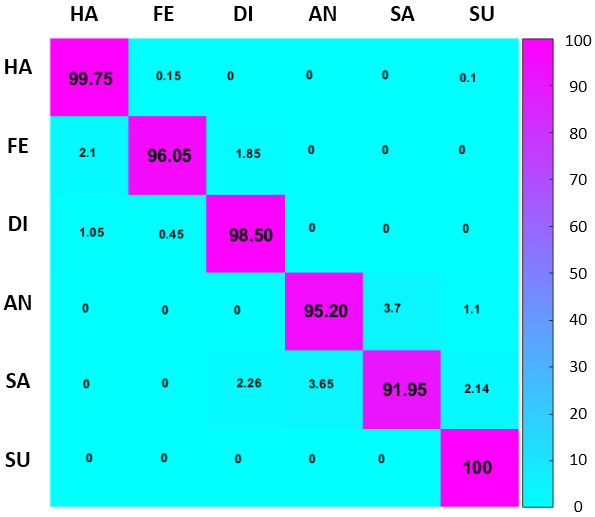}
\caption{Confusion matrix of BU-3DFE dataset.}
\label{fig:conf_BU}
\end{figure}

To assess the proposed system, a 10 fold-cross validation protocol is employed. 90 subjects of the BU-3DFE dataset are then used for training, where 10 subjects are used for the test. Results are averaged across the ten-folds and displayed in the following sections. Fig.~\ref{fig:rates_BU_BO} (A) shows the classification rate of each expression.
From Fig.~\ref{fig:rates_BU_BO} (A), it is clear that the proposed method recognizes better the expressions of happiness and surprise. Anger and Sad expressions have the lowest performance by 95.20\% and 91.95\% respectively. These performances are explained by the fact that distinguishing between these two expressions is a difficult task, which explains their confusion as presented in Fig.~\ref{fig:conf_BU}. It should be noted that these results are obtained using the best system setting according to the codebook and deep covariance matrices size. More details about the system setting and the effect of the codebook sizes can be found in the following sections.

\begin{figure}[h]
\centering
\includegraphics[width=.28\textwidth]{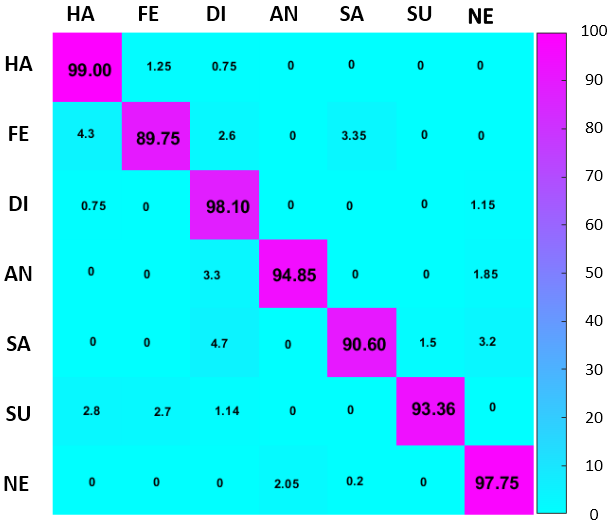}
\caption{Confusion matrix of Bosphorus dataset.}
\label{fig:conf_BOS}
\end{figure}

\subsubsection{Performances on Bosphorus dataset}
To evaluate the proposed system on the Bosphorus dataset, we follow the standard protocol (10 fold-cross validation) as in prior methods \cite{azazi2015towards}. Classification performances are shown in Fig.~\ref{fig:rates_BU_BO} (B). With this dataset, it is clear that happiness and disgust are the best recognized expressions with classification rates of 99.00\% and 98.10\% respectively. Fear and sad expressions are more challenging and give 89.75\% and 90.60\% respectively. thus, distinguishing the subjects of Bosphorus dataset with fear and surprise expressions is a very hard task, which explains their confusion as presented in Fig.~\ref{fig:conf_BOS}. It should also be noted that these results are obtained using the best system setting.


\begin{table*}[p]
\centering
\small
\caption{Comparison between the performance of the proposed method and state-of-the-art methods on the BU-3DFE dataset.}
\begin{tabular}{l c c c c}\hline \hline
\textbf{Method}&\textbf{Images}&\textbf{Landmark}&\textbf{Classifier} &\textbf{Rate}\\ \hline  \hline
Zhen et al. 2016 \cite{zhen2016muscular}&   3D mesh       &      automatic      &      SVM+HMM  &83.20\%      \\ 
Jan et al. 2018 \cite{jan2018accurate} &   3D depth+texture      &   automatic          &     SVM &88.54\%   \\
Li et al. 2017 \cite{li2017multimodal} &   2D+3D  &    automatic    &  SVM &86.86\%   \\
Vo et al. 2019 \cite{vo20193d}&2D+3D& automatic&softmax&84.30\%    \\ 
Huynh et al 2016 \cite{huynh2016convolutional} & 2D+3D        &       -               &     CNN  &92.73\%         \\
Derkach et al. 2017 \cite{derkach2017local}&       3D         &     86 manual    &     SVM &81.03\%      \\ 
Jiao et al. 2020 \cite{jiao20202d+} &2D+3D& automatic&-&89.72\% \\ 
Wei et al. 2018 \cite{wei2018unsupervised} &2D+3D&30 automatic&SVM&88.03\% \\ 
Ly et al. 2019 \cite{ly2019novel}&2D+3D&automatic&SVM&87.66\% \\ 
Shao et al. 2019 \cite{shao2019three}&2D+3D&automatic&Softmax&86.50\% \\ 
Fu et al. 2019 \cite{fu2019ferlrtc}&2D+3D& automatic&SVM&82.89\%    \\ \hline \hline

&  \textit{3D mesh / 2D depth map}&& & \\ \hline \hline

\textbf{Our (Shallow COV)}        & 3D mesh     & 40 automatic     &     SVM  &90.50\%    \\ 
\textbf{Our (Deep VGG-16 COV)}&     2D depth     &     automatic    &  SVM &94.50\%   \\ 
\textbf{Our (Deep AlexNet COV)}&     2D depth     &     automatic    &  SVM &94.20\%   \\ 
\textbf{Our (Deep VGG-16 COV + Deep AlexNet COV)}&     2D depth     &     automatic    &  SVM &94.99\%   \\ 
\textbf{Our(Deep VGG-16 + Shallow COV)}         & 3D mesh+2D depth     &  automatic     &     SVM  &96.73\%    \\ 
\textbf{Our(Deep AlexNet + Shallow COV)}         & 3D mesh+2D depth     &  automatic     &     SVM  &96.15\%    \\ 
\textbf{Our(Deep AlexNet VGG-16 + Shallow COV)}         & 3D mesh+2D depth     &  automatic     &     SVM  &\textbf{96.90\%}    \\ \hline \hline

&  \textit{3D mesh / Principal curvatures map}&& & \\ \hline \hline

\textbf{Our (Deep VGG-16 COV)}&     Principal curvature     &     automatic    &  SVM &  92.14\% \\ 
\textbf{Our (Deep AlexNet COV)}&     Principal curvature     &     automatic    &  SVM &92.00\%   \\ 
\textbf{Our (Deep VGG-16 COV + Deep AlexNet COV)}&     Principal curvature     &     automatic    &  SVM &  93.00\% \\ 
\textbf{Our(Deep VGG-16 + Shallow COV)}         & 3D mesh+Principal curvature     &  automatic     &     SVM  &94.71\%   \\ 
\textbf{Our(Deep AlexNet + Shallow COV)}         & 3D mesh+Principal curvature     &  automatic     &     SVM  &94.32\%    \\ 
\textbf{Our(Deep VGG-16 + AlexNet + Shallow COV)}         & 3D mesh+Principal curvature     &  automatic     &     SVM  &95.15\%   \\ \hline \hline

\end{tabular}
\label{tab:compare2}
\end{table*}

\begin{table*}[p]
\centering
\small
\caption{Comparison between the performance of the proposed method and state-of-the-art methods on the Bosphorus dataset.}
\begin{tabular}{l c c c c}\hline \hline
\textbf{Method}&\textbf{Images}&\textbf{Landmark}&\textbf{Classifier} &\textbf{Rate}\\ \hline  \hline
Ramya et al.2020 \cite{ramya20203d}&3D&automatic&SVM&87.69\%    \\                    
Wang et al. 2013 \cite{wang2013learning}    & 3D     &    automatic       &     SVM  &76.56\%  \\ 
Jiao et al. 2020 \cite{jiao20202d+}        & 2D+3D  &    automatic       &     -   &83.63\%        \\ 
Vo et al. 2019 \cite{vo20193d}&2D+3D& automatic&softmax&82.40\%    \\ 

Fu et al. 2019 \cite{fu2019ferlrtc}&2D+3D& automatic&SVM&75.93\%    \\ 

Wei et al. 2018 \cite{wei2018unsupervised}&2D+3D&30 automatic&SVM&82.50\% \\ \hline \hline

&  \textit{3D mesh / 2D depth map}&& & \\ \hline \hline

\textbf{Our (Shallow COV)}        & 3D mesh     & 40 automatic     &     SVM  &86.17\%    \\ 
\textbf{Our (Deep VGG-16 COV)}        & 2D depth     &  automatic     &     SVM  &92.20\%    \\ 
\textbf{Our (Deep AlexNet COV)}        & 2D depth     &  automatic     &     SVM  &93.30\%    \\ 
\textbf{Our (Deep VGG-16 COV + Deep AlexNet COV)}        & 2D depth     &  automatic     &     SVM  &93.95\%    \\ 
\textbf{Our(Deep VGG-16 + Shallow COV)}         & 3D mesh+2D depth     &  automatic     &     SVM  &94.26\%    \\ 
\textbf{Our(Deep AlexNet + Shallow COV)}         & 3D mesh+2D depth     &  automatic     &     SVM  &94.55\%    \\ 
\textbf{Our(Deep VGG-16 + AlexNet + Shallow COV)}         & 3D mesh+2D depth     &  automatic     &     SVM  &\textbf{94.77\%}    \\ \hline \hline

&  \textit{3D mesh / Principal curvatures map}&& & \\ \hline \hline

\textbf{Our (Deep VGG-16 COV)}        & Principal curvature     &  automatic     &     SVM  &91.16\%    \\ 
\textbf{Our (Deep AlexNet COV)}        & Principal curvature     &  automatic     &     SVM  &91.50\%    \\ 
\textbf{Our (Deep VGG-16 COV + Deep AlexNet COV)}        & Principal curvature     &  automatic     &     SVM  &92.50\%    \\ 
\textbf{Our(Deep VGG-16 + Shallow COV)}         & 3D mesh+Principal curvature     &  automatic     &     SVM  &93.88\%    \\ 
\textbf{Our(Deep AlexNet + Shallow COV)}         & 3D mesh+Principal curvature     &  automatic     &     SVM  &93.95\%   \\ 
\textbf{Our(Deep VGG-16 + AlexNet + Shallow COV)}         & 3D mesh+Principal curvature     &  automatic     &     SVM  &94.25\%    \\ \hline \hline

\end{tabular}
\label{tab:compare3}
\end{table*}

\subsection{Results comparison and analysis}
\subsubsection{BU-3DFE}
Table~\ref{tab:compare2} presents the performance comparison of the proposed method with those of the literature on the BU-3DFE dataset. The reported results are obtained using the same setting of evaluation. When dealing with the whole dataset, Jan et~al. 2018 \cite{jan2018accurate} achieved 88.54\% using a deep CNN model with hand-crafted features. Derkach et al. 2017 \cite{derkach2017local} used graph laplacian features and obtained an accuracy of 81.03\%. In \cite{wei2018unsupervised}, the authors applied Wasserstein distance with transformed training strategy and obtained 88.03\%. Ly et~al. 2019 \cite{ly2019novel} applied deep 2D and 3D multi-modal approach and SVM classifier. They achieved 87.66\%. Finally, Shao et~al. 2019 \cite{shao2019three} have conducted three pre-trained CNNs (i.e. shallow network, dual-branch CNN and CNN with transfer learning technique) only on 5 expressions and obtained 86.50\%. 

The proposed method overcomes the previous methods and achieved 96.90\% using the quantization of the combination VGG-16, AlexNet, and Shallow covariance features. 

Table \ref{tab:compare_BU} shows a comparison between the obtained classification rates and state-of-the-art ones that aim to recognize the six prototypical expressions. It is clear that the proposed method achieved the best classification rate with FE, DI, AN, SA and SU expressions using the combination of shallow and deep based features by (96.05\%, 98.50\%, 95.20\%, 91.95\% and 100\%) respectively. With HA expression, Huynh et~al. 2016 \cite{huynh2016convolutional} obtained 100\% as highest rate.

\begin{table*}[h]         
\centering
\tiny
\caption{Comparison of classification rates (\%) per expression of our proposed method (shallow features only + best combination) with state-of-the-art methods on the BU-3DFE dataset.}
\tiny
\resizebox{\textwidth}{!}{
\begin{tabular}{l c c c c c c c}\hline 
\textbf{Method}&\textbf{HA}&\textbf{FE}&\textbf{SA}&\textbf{AN}&\textbf{DI}&\textbf{SU}&\textbf{Average} \\ \hline  
Zhen et al. 2016 \cite{zhen2016muscular} &94.6&63.3& 79.2&79.5&85.7& 96.1&83.2\\      
Berretti et al. 2010 \cite{berretti2010set} & 86.9&63.6& 64.6 &81.7&73.6&94.8& 77.53  \\  
Li et al. 2017 \cite{li2017multimodal} &96.26 & 79.24&81.18 &82.08 &84.94 &97.43 & 86.86 \\  
Huynh et al. 2016 \cite{huynh2016convolutional} &\textbf{100}&86.7& 87.5&91.3&95.2&95.7&92.73 \\ 
Derkach et al. 2017 \cite{derkach2017local}&89.50&65.12&  77.20&85.58&75.31&93.50&81.03 \\  
Lemaire et al. 2013 \cite{lemaire2013fully}&  89.8  &  64.6  & 74.5 & 74.1  & 74.9  & 90.9  &  78.13 \\ 
Vo et al. 2019 \cite{vo20193d}  &  87.50  &  66.88 & 81.25 & 80.00  & 79.06  & 91.25  &  84.30 \\ 
Fu et al. 2019 \cite{fu2019ferlrtc}& 92.25 &70.75 & 78.91 & 80.92  & 78.67  & 95.83  &  82.89 \\ 


 \textbf{Our method (Shallow only)}&95.5&89.67&  83.33&86.00&92.37&96.33&90.50\\

 \textbf{Our method (Best)}&99.75&\textbf{96.05}& \textbf{91.95} &\textbf{95.20}&  \textbf{98.50} &\textbf{100}&\textbf{96.90}\\ \hline 

\end{tabular}
}
\label{tab:compare_BU}
\end{table*}
\begin{table*}[h]
\centering
\tiny
\caption{Comparison of classification rates (\%) per expression of our proposed method (shallow features only + best combination) with state-of-the-art methods on the Bosphorus dataset.}
\resizebox{\textwidth}{!}{
\begin{tabular}{l c c c c c c c c}\hline 
\textbf{Method}&\textbf{HA}&\textbf{FE}&\textbf{SA}&\textbf{AN}&\textbf{DI}&\textbf{SU}&\textbf{NE}  &\textbf{Average} \\ \hline  
Wang et al. 2013 \cite{wang2013learning}& 92.50   & 62.80  &  74.50 & 63.50  &  70.60 &\textbf{95.60}&  - &76.56   \\               
Fu et al. 2019 \cite{fu2019ferlrtc}    & 92.97  & 63.83  & 65.97 &  77.37 & 67.03  & 88.40  & -  & 75.95\\
Ramya et al. 2020 \cite{ramya20203d}  &83.08 & \textbf{96.92}  & 78.46  &  87.69  & 87.69  &  93.85  &  86.15   &  87.69  \\ 
Azazi et al. 2015 \cite{azazi2015towards}&97.50&86.25&67.50&  82.50 & 90.00  & 83.75  & 81.25   & 84.10\\  
\textbf{Our method (Shallow only)}& 93.00  & 81.00  & 79.75  &86.25&85.25&90.50&87.50&86.17\\
\textbf{Our method (Best)}&\textbf{99.00} & 89.75  &  \textbf{90.60}  &\textbf{94.85}& \textbf{98.10}&93.36&\textbf{97.75}&\textbf{94.77}\\ \hline
\end{tabular}
}
\label{tab:compare_BS}
\end{table*}

\subsubsection{Bosphorus}
Table~\ref{tab:compare3} presents the performance comparison between the proposed method with those of the literature on the Bosphorus dataset. It is clear that the proposed method gives the best accuracy (94.77\%) using the quantization of the combination VGG-16, AlexNet, and Shallow covariance features. The use of deep covariance features (VGG-16 and Alexnet) separately with shallow ones achieves a slightly lower recognition performance by 94.26\% and 94.55\% respectively. This combination has improved the previous one obtained by each deep model separately by 92.20\% and 92.30\%. 

The proposed method outperforms prior methods. For example, Ramya et~al. 2020 \cite{ramya20203d} applied a transfer learning based-technique to fine-tune the pre-trained model AlexNet after computing local binary pattern (LBP) and local bidirectional pattern features. They achieved 87.69\%. Also, Vo et~al. 2019 \cite{vo20193d} has achieved 82.40 using multi-view CNN. In \cite{chun2013facial} the authors used 2D transformed images and the texture information, they obtained 76.98\%. Finally, Jiao et~al. 2020 \cite{jiao20202d+} applied VGG-16 and a trained network from scratch and obtained 82.50\%.

Table \ref{tab:compare_BS} shows a comparison between the obtained classification rates and state-of-the-art ones in order to recognize the six prototypical expressions and also the neutral face. The reported results show that Ramya et~al. 2020 \cite{ramya20203d} outperformed the other methods with FE expression by 96.92\%. When dealing with SU expression, Wang et~al. 2013 \cite{wang2013learning} has achieved the highest classification rate by 95.60\%. Our proposed method on the other hand outperformed the state-of-the-art methods when dealing with the expressions: HA, DI, AN, SA and NE by 99.00\%, 98.10\%, 94.85\%, 90.60\% and 95.75\% respectively.

\subsection{Effect of the codebook size}
To further evaluate the performance of the proposed method, we study in this section the effect of the codebook size on the classification rate. We evaluated 7 sizes (i.e. 16, 32, 64, 128, 256, 512 and 1024). By intuition, if the codebook size is too small, the histogram feature loses discriminant power to classify the expressions, whereas the performance increases when the codebook size grows. 

\begin{figure}[h]
 \centering
\begin{subfigure}{0.47\textwidth}
\includegraphics[width=.99\textwidth]{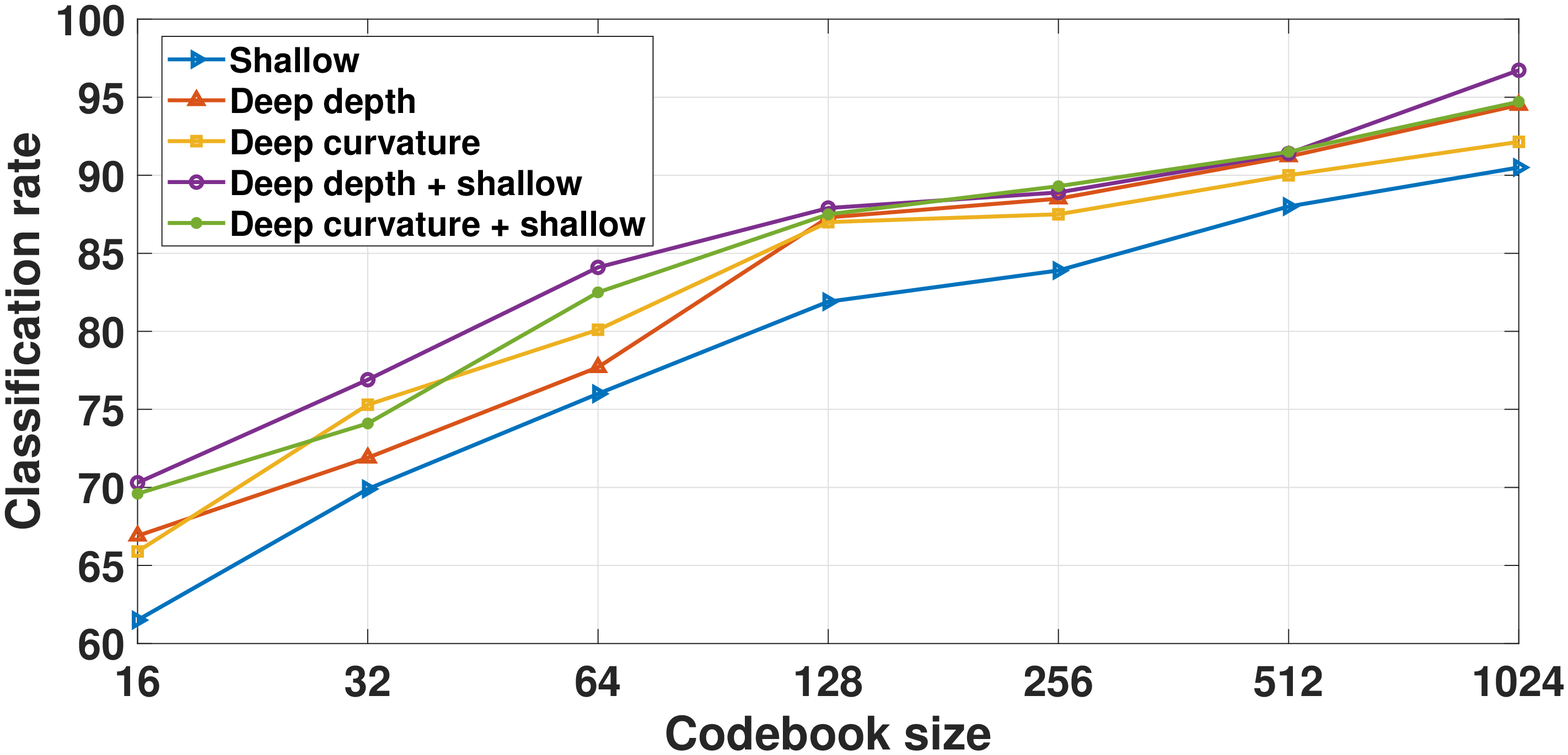}
\subcaption{VGG-16}\label{fig:nnn1}
\end{subfigure}
~
\begin{subfigure}{0.50\textwidth}
\includegraphics[width=.99\textwidth]{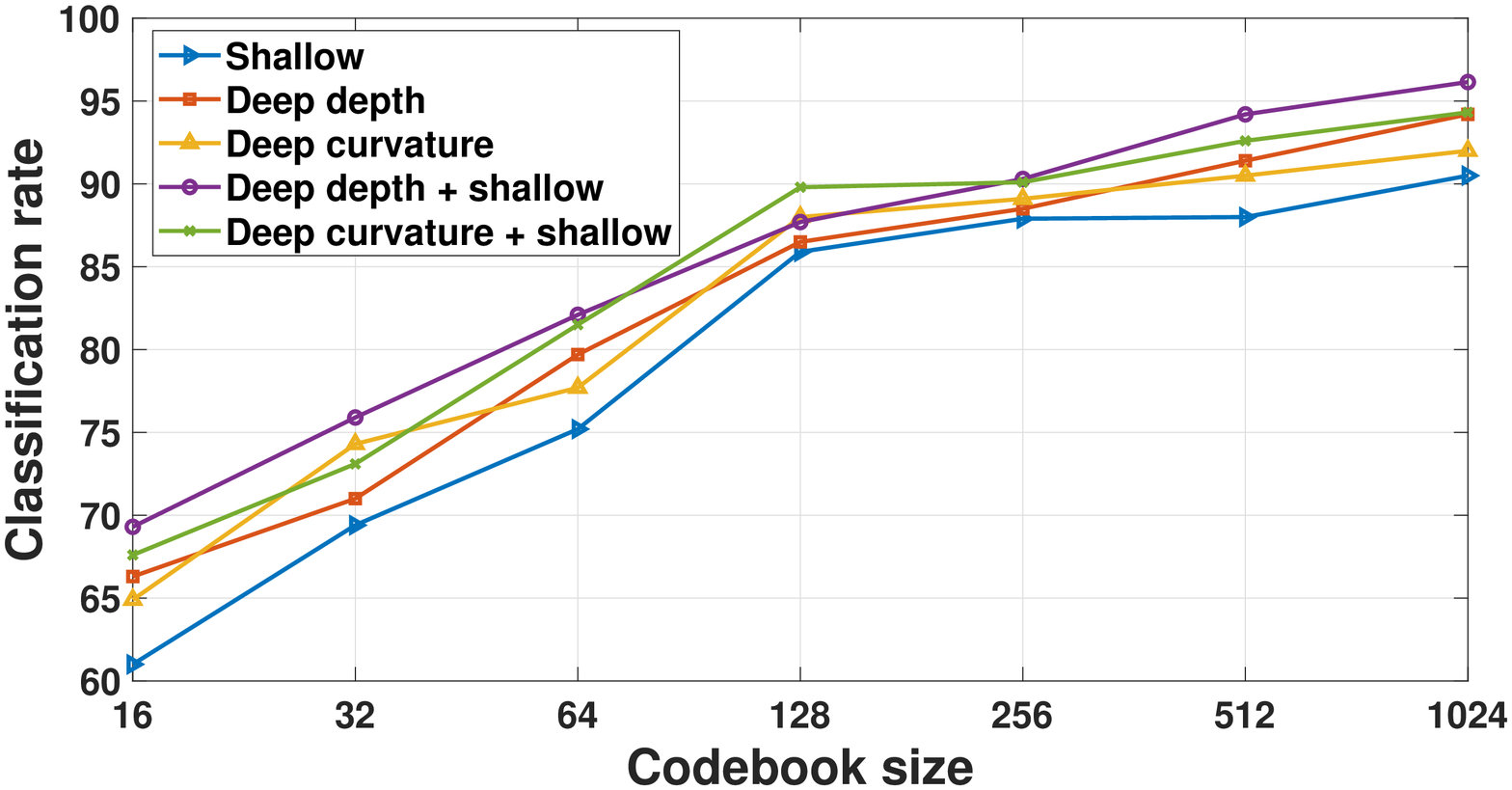}
\subcaption{AlexNet}\label{fig:nnn2}
\end{subfigure}

\caption{Effect of the codebook size on the classification performance. The reported results are obtained on the BU-3DFE dataset using the deep codebooks of VGG-16 and AlexNet separately, each deep codebook is combined with the shallow codebook. The displayed sizes are the same for each codebook.}
\label{fig:codebook_BU_VGG22}
\end{figure}

Fig.~\ref{fig:nnn1} presents the improvement of the classification rate according to the codebook size using the shallow and VGG-16 co-varied features on BU-3DFE dataset. The best classification rate is achieved by the combination of (Deep depth + shallow) codebooks, followed by (Deep depth + shallow), (Deep depth), and (Deep curvature) codebooks respectively. As expected, the shallow codebook gives the lowest classification rate since quantized covariance descriptors don't capture deeper features compared to VGG-16 ones. Nevertheless, integrating a shallow codebook with a deep codebook improves the performance of the proposed method. Overall, the classification rate grows when the codebook size gets bigger and almost stabilizes from the size 512. 

Fig.~\ref{fig:nnn2} presents the variation of the classification rate according to the codebook size using the shallow and AlexNet co-varied features. The same discipline of the previous figure is maintaining here, where the combination (Deep depth + shallow) gives the highest performance.

\begin{figure}[h]
 \centering
\begin{subfigure}{0.47\textwidth}
\includegraphics[width=.99\textwidth]{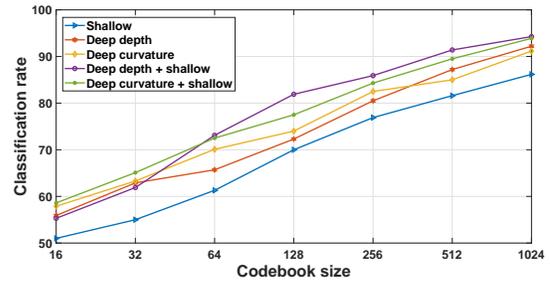}
\subcaption{VGG-16}\label{fig:nnn3}
\end{subfigure}
~
\begin{subfigure}{0.50\textwidth}
\includegraphics[width=.99\textwidth]{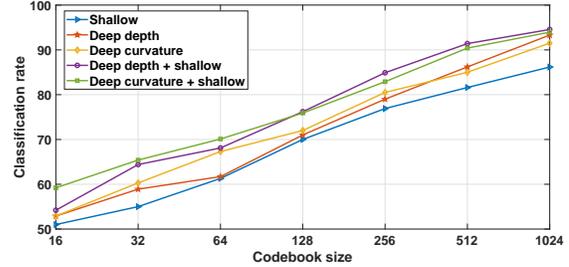}
\subcaption{AlexNet}\label{fig:nnn4}
\end{subfigure}

\caption{Effect of the codebook size on the FER performance. The displayed results are obtained on the Bosphorus dataset using the deep codebooks of VGG-16 and AlexNet separately, each deep codebook is combined with the shallow codebook. The displayed sizes are the same for each codebook.}
\label{fig:codebook_Bos_VGG22}
\end{figure}

The same study has been conducted on the Bosphorus dataset. From Fig.~\ref{fig:nnn3} and Fig.~\ref{fig:nnn4}, we can notice that the shallow codebook gives the lowest classification rate as shown before on BU-3DFE. When dealing with deep-based codebook, curvature-based codebook initially outperforms depth-based codebook with small codebook size (i.e. from 16 to 256). In contrast, using grand codebook size (i.e. 512 and 1024), the depth-based codebook becomes better and slightly outperforms the curvature-based one. This discipline can be explained by the fact that depth information needs a grand quantized feature to be sufficiently encoded.

\begin{figure*}[h]
 \centering
\begin{subfigure}{0.47\textwidth}
\includegraphics[width=.98\textwidth]{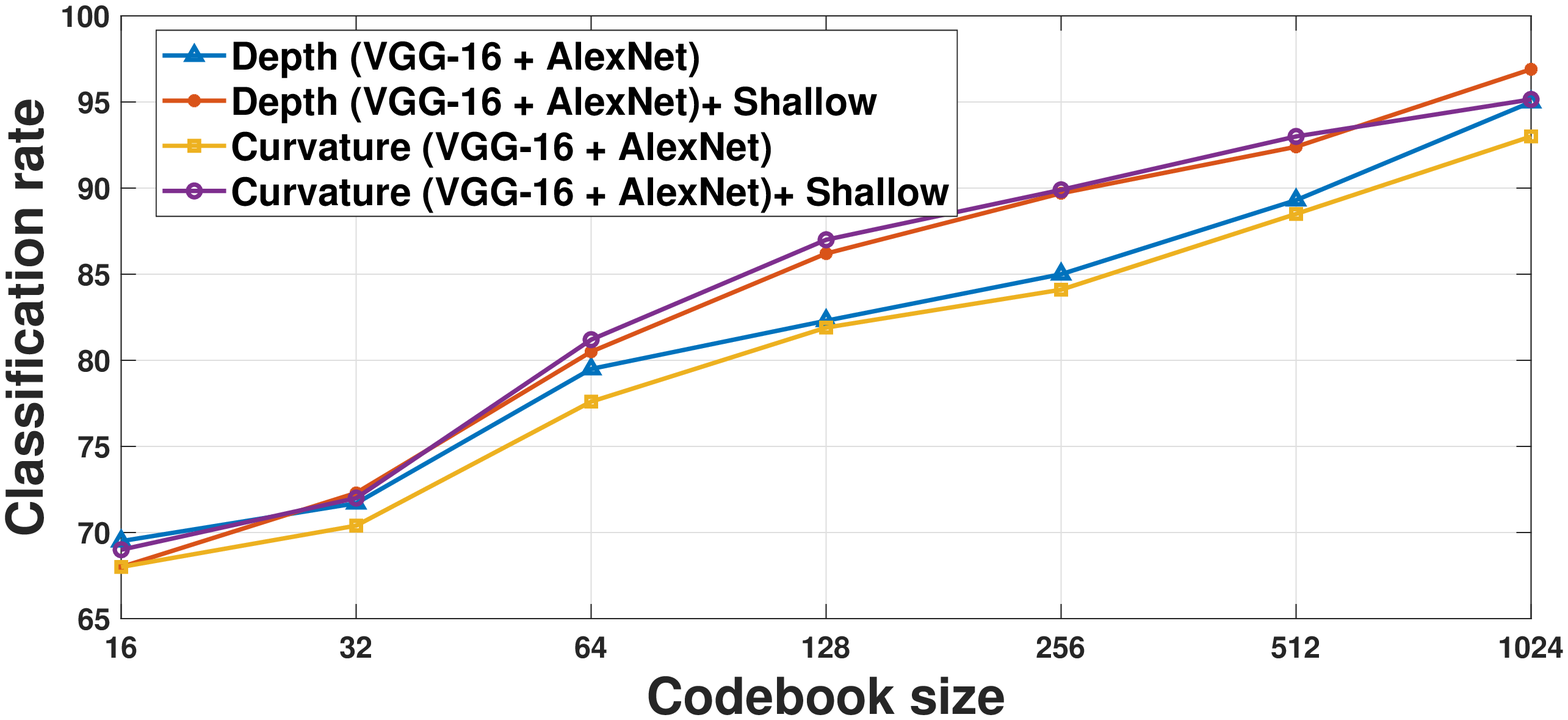}
\subcaption{BU-3DFE}\label{fig:nnn5}
\end{subfigure}
~
\begin{subfigure}{0.49\textwidth}
\includegraphics[width=.99\textwidth]{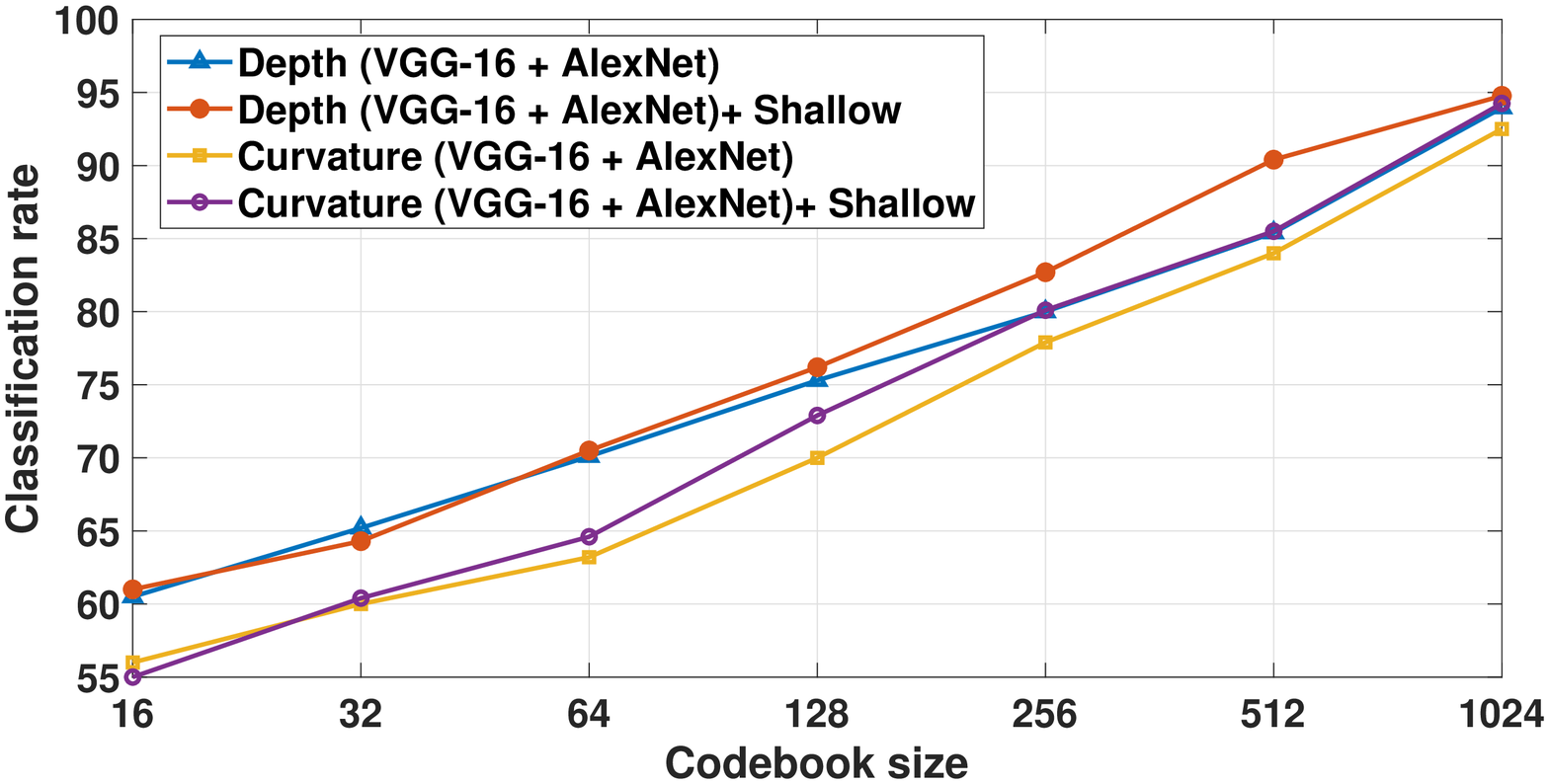}
\subcaption{Bosphorus}\label{fig:nnn6}
\end{subfigure}
\caption{Effect of the codebook size on the FER performance on the Bosphorus and BU-3DFE datasets. The deep codebooks of VGG-16 and AlexNet are combined with each other, and with the shallow codebook as well.}
\label{fig:codebook_BU_VGG_AlexNet}
\end{figure*}

The performance of the combination between VGG-16 and AlexNet co-varied features are presented in the Fig.~\ref{fig:codebook_BU_VGG_AlexNet} according to the codebook size. Using the two datasets, the highest performance is realized by the combination \textit{Depth (VGG-16 + AlexNet)+ Shallow} and \textit{Curvature (VGG-16 + AlexNet)+ Shallow} codebooks respectively. This achievement can be explicated by the fact that VGG-16-based and AlexNet codebooks are complimentary. Moreover, the shallow-based codebook enhances the performance of the proposed method through the face's encoded geometry and topological information. The lowest performance, in contrast, is obtained using the combination \textit{Curvature (VGG-16 + AlexNet)}. Thus, deeper features extracted from curvature map images may lack some geometrical information that can be provided by the shallow features. This further demonstrates that the combination of the two codebooks (shallow and deep ones) provides high discriminative power and thus is suitable for multi-class SVM classification to overcome the huge challenges in the 3D FER task.

\section{Discussion}
\label{sec:disc}
The reported results above show that our proposed method achieved state-of-the-art performance without using additional training data or facial registration. The reliable accuracy is obtained thanks to the high discrimination power of the SVM classifier. This high-performance power cannot be achieved without using proper discriminative face descriptors. Basically, deep and shallow features-based covariance descriptors capture all the information of the facial expressions and their correlation. Moreover, we show that deep covariance-based descriptors give higher classification rates comparing to shallow ones. This is a predictable discipline due to the high efficiency of deep features reinforced by the covariance matrices learning through the BiMap layer and the non-linear transformation. Since covariance matrices inherit their performance from the used features, CNN models are designed to extract the best features from the 2D transformed face images (i.e. depth and curvature maps). Shallow features on the other hand extract more details about geometrical and spatial information. More particularly, the reported results show that the pre-trained deep features on depth map images outperformed the curvature-based one when dealing with BU-3DFE and Bosphorus datasets. This difference is obtained due to the higher description of the face structure insured by the 2D depth map images. 

We can also notice that incorporating VGG-16 and AlexNet models boosts the performance of the proposed framework compared to the use of each model separately. This is because the two models have different architectures. Thereby, the convolution layer used to generate our deep features are relatively different; each of them encodes different feature abstraction levels that could capture complementary characteristics from the facial expression images. Embedding additional pre-trained models are supposed to improve the performance of the proposed framework, however, the choice of the model should be thoroughly studied according to the number of layers and the degree of similarity between the trained database used to initialize these models to be used for the FER task.

The obtained experimental results further demonstrate the capacity to notably improve the FER performance using pre-trained deep network structures combined with a shallow structure. This combination through the two codebooks could overcome the shortage of training data and over-fitting problem. We can conclude that these two covariance-based methods are complementary and their quantization using the BoF paradigm is suitable for the 3D FER task. It's worth noting that this complementarity is achieved due to the 2D+3D multi-modality, where each modality can capture different characteristics of the facial expression. This is an advantage compared to state-of-the-art methods that generally extracted shallow and deep features from the same 2D image modality, for example, Yang et~al. \cite{yang2017facial} extracted deep features using pre-trained models, and LBP features from the same 2D images. When dealing with the same 3D datasets, the fine-tuning applied by Ramya et~al. 2019 of the AlexNet model with additional shallow CNN (called multi-channel framework) doesn't outperform our proposed method using the AlexNet model separately (87.69\% and 93.30\%; respectively). The reported results are the overall accuracies obtained using Bosphoros dataset. Thereby, Our proposed method outperforms Rampya et al.'s method with six expressions from seven. This finding further demonstrates the efficiency of the proposed system through the quantization of the co-varied generated deep features.

\section{Conclusion}
A new methodology for 3D FER is presented. To deal with the significant variations from facial expression images, we have employed covariance matrices that ensure an efficient combination of different features extracted from distinct image modalities. To do so, we have adopted two different types of features to extract the covariance matrices (i.e. deep and shallow features). The purpose is to capture not only the best features extracted from the last convolutional layer of VGG-16 and AlexNet models but also to capture more geometrical and spatial details from the 3D expressive faces. Note that our method is generic so that other pre-trained deep learning models can be added (e.g. ResNet50), it can also be computed from other 2D transformed images such as Shape Index or Curvedness maps. Covariance matrices, however, belong to the Riemannian structure of $Sym_d^+$ space. Therefore, classifying facial expressions using their covariance-based descriptors needs a global and optimal description to be able to employ traditional classification algorithms. Therefore, we first apply linear and non-linear transformations using deep covariance matrices learning in order to reduce their size and enhance their discrimination power while preserving their manifold structure. Second, we flatten the obtained covariance matrices (deep and shallow ones). The BoF paradigm is then applied to the two sets of the obtained features of the flattened covariance matrices. A multi-class SVM is finally employed to classify the expressions after being trained using the global quantized feature vector (i.e. deep and shallow codebooks). The displayed performances obtained on the BU-3DFE and Bosphorus datasets demonstrate that the deep codebook gives higher classification rates comparing to the shallow codebook. Furthermore, their combination outperformed the deep codebook single-handed. This discipline proves that the two codebook-based methods are complementary and efficiently classify 3D facial expressions. Moreover, the reported results also confirm the advantage of the application of the BoF paradigm to classify facial expressions using 3D data in comparison to state-of-the-art methods. In future work, dynamic features could be added to the proposed system to boost the classification rate of the 3D FER (BU-4DFE dataset). Also, we will investigate the use of point cloud as input in CNN by applying the PointNet architecture proposed in \cite{qi2017pointnet}. We also look at the application of generative models to provide additional training data to deal with the problem of small 3D FER datasets and to further enhance the efficiency of the proposed framework.




%

\ifCLASSOPTIONcompsoc
  \section*{Acknowledgments}
\else
  \section*{Acknowledgment}
\fi

The authors would like to thank the DGRSDT, Algeria.

\ifCLASSOPTIONcaptionsoff
  \newpage
\fi

\bibliographystyle{IEEEtran}
\bibliography{refs}




\end{document}